\documentclass[letterpaper]{article} 
\usepackage{aaai25}  
\nocopyright
\usepackage{times}  
\usepackage{helvet}  
\usepackage{courier}  
\usepackage[hyphens]{url}  
\usepackage{graphicx} 
\usepackage{natbib}  
\usepackage{caption} 
\usepackage{amsmath}
\usepackage{booktabs}
\title{Where Language-Model Vocabulary Rose on Stack Exchange}
\author{Fengming Liu}
\affiliations{University of Chicago\\
fengmingliu@uchicago.edu}

\begin{document}
\maketitle

\begin{abstract}
Stack Overflow banned answers generated by ChatGPT five days after the model's release, and many other Stack Exchange sites announced similar rules. I measure the vocabulary of the answers posted afterwards, 3.8 million answers on 170 sites from 2019 to 2024. A fixed list of 56 words that language models overuse rose about 14 percent above its pre-2023 trend, while the highest of 1{,}000 frequency-matched lists of ordinary words rose about 7 percent. In a screen of 15 site traits measured before the release, the strongest predictor of where the rise happened was the share of answers that received a body edit within 30 days of posting. On the least edited third of sites the list rose about 29 percent relative to matched controls, and on the most edited third it did not change. The gradient is also present among answers that were never edited. It follows the share of answers edited by their own authors, and the share edited by others adds little once that share is held fixed. Use of the model is unobserved, so these results cannot separate lower use on sites with a high self-edit share from any effect of editing on what the vocabulary records. Any vocabulary-based estimate for the network as a whole therefore pools sites where the marker vocabulary rose sharply with sites where it did not change.

\end{abstract}

\section{Introduction}

On 5 December 2022, five days after ChatGPT was released, Stack Overflow banned answers generated by it \citep{sepolicy2022}. The moderators gave accuracy as their reason. In their account generated answers read well, were often wrong, and arrived faster than volunteers could check them. Over the following weeks moderators of many other sites in the network announced similar rules. I have no record of which sites adopted a rule or when. Prior work has measured the drop in questions on these sites \citep{delriochanona2024,burtch2024} and has compared the model's answers with human ones \citep{kabir2024}. I look at the answers people went on posting, 3.8 million of them on 170 sites from January 2019 to March 2024.

The excess-vocabulary studies identified words that language models use far more often than people do \citep{kobak2025,liang2024monitoring}, and the outcome here is a fixed list of 56 such words taken from them, set before I read any Stack Exchange text. From the second quarter of 2023 the share of answers containing list words was about 14 percent above its 2019 to 2022 trend across the network. The highest of 1{,}000 lists of ordinary words with the same frequency profile rose about 7 percent. Without its most frequent word, \textit{ensure}, the list still rose 10 percent.

I planned to explain where the rise happened by code exposure, since programming questions were an obvious use of the early model. Code exposure did predict the rise, but it is correlated with other site traits. When I screened 15 traits on equal terms, the strongest predictor was one I had not planned to test. It is the share of a site's answers that received a body edit within 30 days of posting, measured on answers from before the release. On the least edited third of sites the list rose about 29 percent relative to matched controls, and on the most edited third it did not change. The gradient is also present among answers that were never edited. It follows the share of answers edited by their own authors, and the share edited by others adds little once that share is held fixed.

I do not observe use of the model. One reading, which I cannot test, is that sites with a high self-edit share used the model less. Under that reading or any other, a vocabulary-based estimate for the whole network averages over sites that differed this much. The variation arose during a period when many sites had rules against generated answers, but without sites known to lack a rule the data say nothing about what the rules did.

\section{Related Work}

\citet{liang2024monitoring} estimated the share of AI-modified peer reviews from shifts in adjective frequencies and later applied the method to consumer complaints, press releases and job postings \citep{liang2025widespread}. \citet{kobak2025} found that excess vocabulary in 2024 biomedical abstracts exceeded anything seen during the pandemic, and \citet{geng2024} found similar shifts in arXiv abstracts. Each of these studies extrapolates a counterfactual for one large corpus whose topics drift slowly. None has a dose that varies across units, and none looks at a place where model text was forbidden.

Stack Overflow activity fell sharply after ChatGPT relative to comparable forums \citep{delriochanona2024}, most where the model could answer the questions the community answered \citep{burtch2024}, and in this work the model is a competitor for questions. When \citet{kabir2024} asked users to compare ChatGPT answers with accepted human answers to the same questions, the model's answers were longer and often wrong, and users still preferred them about a third of the time, usually for their style. I measure the style of the answers that people continued to post.

In earlier work on embedding homogeneity \citep{anon2026methods} I showed that a popular cosine measure on Reddit and Stack Exchange tracks text length, and that an apparent post-ChatGPT convergence of technical sites came from their answers growing longer.

\section{Data and Design}

\paragraph{Corpus.} I use the April 2024 Stack Exchange data dump and keep every main site with at least 500 answers between January 2019 and March 2024 and at least 100 in the exposure window. Stack Overflow would dominate every weighted estimate and its collapse after ChatGPT is well documented, so it is excluded along with the meta sites. That leaves 170 sites and 3.83 million answers. From each answer I strip code blocks, inline code, HTML tags and URLs and keep the prose.

\paragraph{Marker list.} The 56 words come from the excess-vocabulary studies and include \textit{delve}, \textit{crucial}, \textit{ensure}, \textit{furthermore}, \textit{showcase} and \textit{tapestry}. Of these, 54 occur in the corpus. For each answer I count the distinct list words per hundred prose tokens, with the denominator floored at 20, and record whether any list word appears. I call the first the marker rate. In 2021 and 2022 the mean marker rate was 0.035 list types per hundred tokens, and 5.6 percent of answers contained at least one list word. For site-level work I aggregate to site quarters, where the marker score is the mean over list words of the share of answers containing the word.

\paragraph{Matched control vocabulary.} General drift could lift any set of words, so I built control lists from the 5{,}546 lowercase words of four or more letters that are absent from the marker list and appear on at least 50 sites. A control list replaces every marker word with a random word from the same twentieth of the pre-period frequency distribution. Drawn many times, these lists give a null distribution for any statistic. Averaged 200 at a time, they give one control score per site quarter.

\paragraph{Site traits.} The planned dose was code exposure, the share of a site's questions from January 2021 to November 2022 that contain a \texttt{<pre>} block. It runs from zero on Anime, Beer and Parenting to 1.00 on Code Review, with a median of 0.03. Fourteen further traits come from the same window, namely mathematical notation, code in answers, answer length, its spread, question length, answer volume, the number of questions, answers per question, the use of lists, headers, summaries and phrases such as \textit{here's}, the pre-period rate of \textit{delve}, and the edit share. Every trait enters as the site's percentile rank centered at zero, so a coefficient approximates the gap between the highest- and the lowest-ranked site.

\paragraph{Edit share.} The dump marks an answer as edited if anyone revised it at any time up to April 2024, so the flag on a 2021 answer can reflect an edit made in 2023. To measure editing before the release, I read the revision history of every answer posted from 2019 on and date each body edit, counting rollbacks of the body as edits. The edit share is the share of answers posted from January 2021 to October 2022 that received a body edit within 30 days of posting, so every edit it counts was made before 1 December 2022. It replaces the dump's flag in the screen and in every estimate below. Its median is 0.32 and its interquartile range 0.24 to 0.39. It ranks sites almost as the dump's flag does, since only 1.7 percent of ever-edited answers from 2021 and 2022 received their first edit after November 2022. An edit counts as a self-edit when the editor owns the answer and as an other-edit otherwise. On the median site 27 percent of answers have a self-edit within 30 days and 6 percent an other-edit.

\paragraph{Specification.} For site $s$ in month $m$,
\begin{equation}
y_{sm} = \alpha_s + \delta_m + \beta\, r_s \cdot \mathbf{1}[m \geq \text{2022-12}] + \lambda_s m + \varepsilon_{sm},
\end{equation}
where $r_s$ is a centered trait rank and $\lambda_s m$ an optional site-specific linear trend. Observations are weighted by the number of answers, capped at 2{,}000, and standard errors are clustered by site \citep{cameron2015}. The month effects absorb any change common to all sites, including the Stack Overflow ban, the May 2023 instruction to moderators to stop relying on AI detectors, and the moderator strike from June to August 2023. Rules adopted by individual sites on different dates are not absorbed. Identifying $\beta$ requires parallel trends across trait levels \citep{callaway2021}, which I probe with a placebo break at December 2021 on data before December 2022, with half-year event studies, and with randomization inference that reassigns the trait across sites 2{,}000 times. For the site-level screen I take, for each site, the log change in its marker score from 2021Q1 to 2022Q3 to the period from 2023Q2 to 2024Q1 and subtract the same log change in its control score. The screen uses the 152 sites with at least 20 answers in each window and a nonzero marker score in both.

\section{Results}

\subsection{The marker list rose across the network}

\begin{figure*}[t]
\centering
\includegraphics[width=\textwidth]{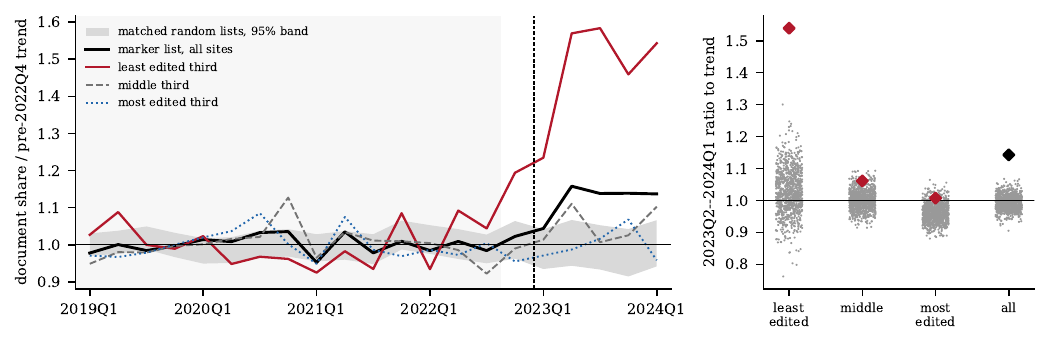}
\caption{Left, marker-list document share relative to a linear trend fitted to 2019Q1--2022Q3, equal weight per site (sites with at least 20 answers in a quarter). The gray band holds 95 percent of 1{,}000 frequency-matched control lists. Colored lines split the sites by tercile of the edit share, the share of 2021--22 answers that received a body edit within 30 days of posting. Right, the 2023Q2--2024Q1 ratio to trend for the marker list (diamond) and for each control list (dots), by edit tercile and for all sites.}
\label{fig:null}
\end{figure*}

For four years the marker score stayed within a few percent of its own linear trend (Figure~\ref{fig:null}). It began to climb in the first quarter of 2023, reached about 14 percent above trend in the second, and remained there in March 2024. Over the same quarters 95 percent of the matched control lists stayed between 9 percent below and 7 percent above their trends, and the highest of the 1{,}000 rose about 7 percent.

A trend fitted to 2021 and 2022 alone, a quadratic over 2019 to 2022 and weighting sites by their answers give rises between 10 and 21 percent, and in every combination the marker list rose more than every control list. A broader turn toward formal writing would lift these words too, so I built two register lists of formal words absent from the marker list. One has 14 connectives such as \textit{however} and \textit{therefore}, the other 20 formal adjectives and adverbs such as \textit{significant} and \textit{appropriate}. With equal site weights both ended 3 to 6 percent below trend under every fit. Weighting by answers narrows the separation, since the connectives then rose up to 10 percent, and for that reason I report the equal-weight figures.

On the least edited third of sites the list ended 54 percent above trend, beyond the highest of its own control lists, while the middle third rose 6 percent and the most edited third 1 percent, both inside the range of their controls (Figure~\ref{fig:null}).

\subsection{Editing predicts where the rise happened}

With site trends the marker rate on the most code-exposed site rose by 0.0082 types per hundred tokens more than on the least exposed, about a quarter of the pre-period mean, so code exposure, the dose I planned, does predict the rise. The event study is flat before the release, and a placebo break a year earlier shows little. The coefficient is significant in weighted specifications without size controls but not unweighted, and with length, size and mathematics in the same regression it falls by more than half and loses significance (Appendix Table~\ref{tab:code}). I then screened all 15 traits on equal terms, regressing each site's marker change net of controls on one trait rank at a time.

\begin{figure}[t]
\centering
\includegraphics[width=\columnwidth]{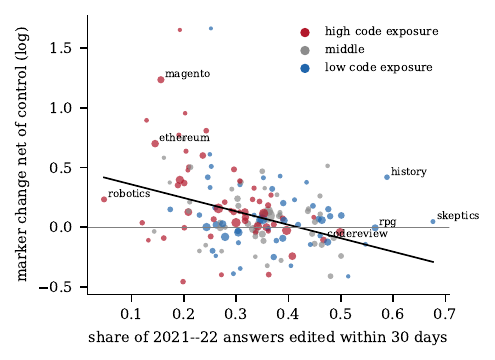}
\caption{Each point is a site. The vertical axis is the log change in marker score from 2021Q1--2022Q3 to 2023Q2--2024Q1 minus the log change in the matched control score. The horizontal axis is the share of 2021--22 answers that received a body edit within 30 days of posting. Point size scales with answer volume and color marks the tercile of code exposure. The line is a weighted least-squares fit.}
\label{fig:edit}
\end{figure}

One at a time, the edit share explains 17 to 18 percent of the variance in the marker change across sites, and no other trait explains more than 4 percent (Table~\ref{tab:screen}). The edit share survives a Bonferroni correction for 15 tests, and it predicts nothing in a placebo window that compares 2019 and 2020 with 2021. By terciles of editing, the least edited third rose 0.26 log points relative to its controls (bootstrap interval 0.16 to 0.38), the middle third 0.08 and the most edited third 0.00. A rise of 0.26 log points is about 29 percent. This figure differs from the 54 percent above because the 54 percent weights sites equally and is measured against each group's own trend, while the 29 percent weights sites by answers and is measured against matched controls on the same sites. The largest risers are small product and technology sites such as Tridion, Magento, Monero and Tezos, where fewer than a quarter of answers received an edit within 30 days (Figure~\ref{fig:edit}). On Skeptics and Role-playing Games more than half did, and their marker scores did not move.

\begin{table}[t]
\centering
\small
\begin{tabular}{lccc}
\toprule
 & Coef.\ ($p$) & $R^2$ & Placebo \\
\midrule
Edit share & $-$0.42 ($<$.001) & .18 & 0.02 \\
Dump's edited flag & $-$0.40 ($<$.001) & .17 & 0.02 \\
Code exposure & 0.15 (.052) & .03 & $-$0.02 \\
\addlinespace
\multicolumn{4}{l}{\textit{Entered together}} \\
\quad edit share & $-$0.40 ($<$.001) & .18 & 0.02 \\
\quad code exposure & 0.06 (.29) & & $-$0.02 \\
\addlinespace
\multicolumn{4}{l}{\textit{With code, size and length}} \\
\quad edit share & $-$0.38 ($<$.001) & .19 & \\
\addlinespace
Self-edit share & $-$0.42 ($<$.001) & .18 & \\
Other-edit share & $-$0.22 (.017) & .05 & \\
\multicolumn{4}{l}{\textit{Self and other together}} \\
\quad self-edit share & $-$0.39 ($<$.001) & .19 & $-$0.01 \\
\quad other-edit share & $-$0.08 (.24) & & 0.05 \\
\addlinespace
Edit share, 2019--20 & $-$0.35 ($<$.001) & .14 & \\
84 largest sites & $-$0.44 ($<$.001) & .24 & $-$0.01 \\
No blockchain sites & $-$0.41 ($<$.001) & .18 & $-$0.01 \\
List without \textit{ensure} & $-$0.27 (.003) & .06 & 0.04 \\
\bottomrule
\end{tabular}
\caption{Site-level screen. Each site's log change in marker score from 2021Q1--2022Q3 to 2023Q2--2024Q1 minus that of its control score, regressed on centered trait ranks, 152 sites, weighted by answers. The placebo compares 2019Q1--2020Q3 with 2021 on 151 sites. The 84 largest sites have at least 1{,}000 answers in the post window (99 sites in their placebo). The blockchain row drops nine blockchain sites, seven of them in the screen, leaving 145 (147 in the placebo). Except for the second row, the edit share counts body edits within 30 days of posting on answers from January 2021 to October 2022.}
\label{tab:screen}
\end{table}

\begin{figure*}[t]
\centering
\includegraphics[width=\textwidth]{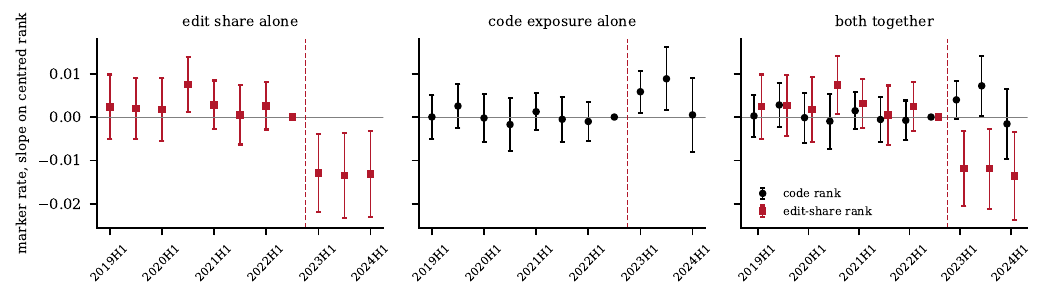}
\caption{Half-year event studies with site and month fixed effects, reference period 2022H2, 95 percent intervals clustered by site. The outcome is the marker rate. Left, on the edit-share rank alone. Middle, on code-exposure rank alone. Right, on both entered together. The dashed line marks the ChatGPT release.}
\label{fig:event}
\end{figure*}

In the monthly panel the post-release change in the marker rate differs by about 0.016 types per hundred tokens from the least to the most edited site, nearly half the pre-period mean of 0.035, with or without site trends (Table~\ref{tab:panel}). None of 2{,}000 random reassignments of the edit share across sites produced a coefficient as large. In the event study the edit gradient stays near zero in every half-year from 2019 to 2022, turns negative in the first half of 2023 and remains so into 2024 (Figure~\ref{fig:event}). Because the reference half-year of the figure includes December 2022, I also fit half-years from June to November and from December to May, referenced on June to November 2022, and there the seven pre-release coefficients are jointly insignificant and the three post-release ones are all negative and significant. The share of answers with any marker is the weaker outcome, since without site trends its placebo is significant and of opposite sign.

A measure built from answers posted in 2019 and 2020, which ranks sites much as the main measure does, gives a similar estimate in the lower rows of both tables, and so do the largest sites alone, the sites outside the blockchain cluster and the list without its most frequent word. In the screen, controlling for each site's pre-period marker score, its ratio to the control score, code exposure, size and length leaves the edit coefficient at $-0.40$. Raising the minimum to 100, 200 or 500 answers per window leaves it between $-0.39$ and $-0.41$. In 500 random splits of the sites into halves, the edit share was the strongest of the 15 traits in the first half 97 percent of the time, and in 98 percent of those cases it was significant with the same sign in the second half.

\begin{table*}[t]
\centering
\small
\setlength{\tabcolsep}{4pt}
\begin{tabular}{lcccc}
\toprule
 & \multicolumn{2}{c}{No site trends} & \multicolumn{2}{c}{Site trends} \\
 & ChatGPT & Placebo & ChatGPT & Placebo \\
\midrule
\multicolumn{5}{l}{\textit{Marker types per 100 tokens}} \\
\quad edit share & $-$0.0155 ($<$.001) & 0.0001 (.96) & $-$0.0165 ($<$.001) & $-$0.0013 (.63) \\
\quad code exposure & 0.0055 (.042) & $-$0.0009 (.64) & 0.0082 (.002) & 0.0014 (.52) \\
\quad edit share, with code & $-$0.0147 ($<$.001) & $-$0.0001 (.98) & $-$0.0151 ($<$.001) & $-$0.0010 (.71) \\
\quad code, with edit share & 0.0035 (.14) & $-$0.0009 (.65) & 0.0061 (.008) & 0.0013 (.56) \\
\quad edit share, with four other traits & $-$0.0148 ($<$.001) & $-$0.0014 (.58) & $-$0.0137 ($<$.001) & $-$0.0010 (.73) \\
\quad code $\times$ edit share & $-$0.0283 (.006) & 0.0043 (.49) & $-$0.0251 (.009) & 0.0095 (.17) \\
\quad self-edit share, with other & $-$0.0167 ($<$.001) & $-$0.0017 (.44) & $-$0.0159 ($<$.001) & $-$0.0029 (.29) \\
\quad other-edit share, with self & 0.0009 (.77) & 0.0051 (.018) & $-$0.0045 (.15) & 0.0036 (.22) \\
\quad edit share, 2019--20 answers & $-$0.0136 ($<$.001) & $-$0.0002 (.92) & $-$0.0143 ($<$.001) & $-$0.0011 (.67) \\
\quad edit share, unedited answers & $-$0.0186 ($<$.001) & $-$0.0003 (.90) & & \\
\quad edit share, edited answers & $-$0.0056 (.15) & 0.0026 (.37) & & \\
\quad edit share, established authors & $-$0.0111 (.003) & $-$0.0010 (.64) & $-$0.0117 (.002) & $-$0.0020 (.56) \\
\quad edit share, newcomers & $-$0.0183 (.002) & 0.0036 (.29) & $-$0.0224 ($<$.001) & $-$0.0001 (.98) \\
\addlinespace
\multicolumn{5}{l}{\textit{Share of answers with any marker}} \\
\quad edit share & $-$0.0139 (.015) & 0.0071 (.040) & $-$0.0256 ($<$.001) & $-$0.0018 (.59) \\
\quad code exposure & 0.0065 (.11) & $-$0.0030 (.29) & 0.0145 ($<$.001) & 0.0051 (.08) \\
\bottomrule
\end{tabular}
\caption{Coefficient on centered trait rank times post, 170 sites, site-month panel with site and month fixed effects, $p$ in parentheses, standard errors clustered by site. Each coefficient approximates the gap between the highest- and the lowest-ranked site. The ChatGPT break is December 2022 and the placebo break December 2021 on data before December 2022. Rows labelled with another trait enter both ranks in one regression, and the four other traits are code exposure, size, length and math. Newcomers are anonymous authors and authors whose first answer on the site came less than a year earlier, and established authors are the rest.}
\label{tab:panel}
\end{table*}

Sites with above-median code exposure and below-median editing rose 0.25 log points relative to controls, high-exposure sites with heavy editing 0.01, and the two low-exposure groups 0.07 and 0.05, so code exposure still matters within the edit split and the interaction of the two ranks in the panel is negative. With the edit share in the regression, code keeps most of its coefficient, significant only with site trends, and in the event study its gradient declines in 2024.

\subsection{The gradient in posted text, self-edits and authors}

Since the dump stores the latest version of each answer, editors could in principle have removed model words from the record. Between the September 2023 and April 2024 copies of the archive fewer than one answer in 150 changed length, and the mean marker count per answer changed by less than 0.0002 types. Among answers that were never edited, whose text is what the author posted, the edit gradient is steeper than in the full sample, and among edited answers it is small and insignificant (Table~\ref{tab:panel}). The never-edited answers are a selected group. The comparison shows that the gradient does not require edits, and it leaves open whether edits add to it. The edit share is also stable within a site, at 0.21 for answers posted in 2019 and in 2024 on the median site of the least edited third and at 0.43 and 0.42 on the most edited third, so the 2023 answers on heavily edited sites were not simply too recent to have been edited.

Self-edits are more than four times as common as other-edits. In the screen each share predicts the rise alone, but entered together the self-edit share keeps its size and the other-edit share falls to about a fifth of it and loses significance (Table~\ref{tab:screen}), so the self-edit share accounts for the gradient. By terciles of the self-edit share the marker list rose 0.28, 0.08 and 0.01 log points relative to controls, and by terciles of the other-edit share 0.18, 0.09 and 0.03. In the panel the other-edit share adds little once the self-edit share is held fixed.

Volume fell more on the least edited sites, where the median site's monthly answers roughly halved, so a change in who posted could in principle produce the gradient. However, the change in the share of answers from newcomers was unrelated to the edit share. The gradient is present among established authors and among newcomers (Table~\ref{tab:panel}), and holding each site's newcomer share at its pre-release level leaves it unchanged.

\subsection{Which words moved}

\begin{figure}[t]
\centering
\includegraphics[width=\columnwidth]{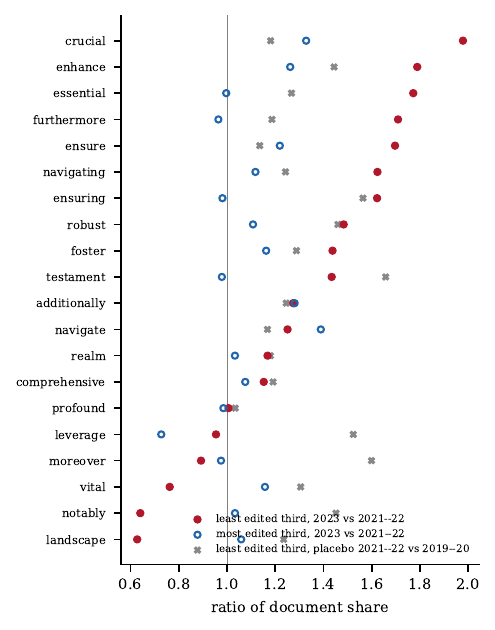}
\caption{Ratio of document share, 2023Q2--2024Q1 over 2021Q1--2022Q3, for list words with a pre-period share of at least 0.2 percent in both the least and the most edited third of sites, by the edit share. Filled circles are the least edited third, open circles the most edited third, crosses the placebo ratio 2021--22 over 2019--20 on the least edited third.}
\label{fig:words}
\end{figure}

\textit{Ensure} alone accounts for about 42 percent of the list's code-exposure coefficient, and without it the network rise is still 10 percent above trend and above every control list. Of the 15 marker words that occur on at least 60 sites, 14 have a negative edit gradient in their own log change and 12 are significant at 0.05, led by \textit{crucial} and \textit{essential}. Dropping any one of these words from the list leaves the edit coefficient in the screen between $-0.40$ and $-0.45$, except \textit{ensure}, which lowers it to $-0.27$. Dropping \textit{additionally}, \textit{crucial}, \textit{ensure}, \textit{essential}, \textit{furthermore} and \textit{moreover} together leaves $-0.39$. On the least edited third of sites the document share of \textit{crucial} nearly doubled and those of \textit{essential}, \textit{furthermore} and \textit{ensure} rose by 70 to 77 percent, while on the most edited third the same words changed by between $-4$ and 33 percent (Figure~\ref{fig:words}). \textit{Delve} appeared in 0.018 percent of pre-period answers and did not rise, and \textit{tapestry}, \textit{showcase} and \textit{intricate} did not rise either.

\subsection{Other outcomes, regular authors and deletion}

ChatGPT answers typically use bulleted lists and headers, and both rose on code-heavy sites, but these estimates lose significance once size, length, math and editing enter, and the header event study shows a pre-trend. Answer length on code-heavy sites shows a borderline post-release rise that continues a contraction beginning in 2019, which the placebo break detects. Mathematical notation, meant as a falsification dose because the 2022 model was poor at mathematics, predicts nothing once size, length and editing are controlled (Appendix Table~\ref{tab:code}).

I define regular authors as the 9{,}573 with at least three answers in total in the 12 months before the release and at least three in the 12 months after. Their marker use rose more on more code-exposed sites, by about 60 percent of the site-level estimates and significantly only for the share of answers with any marker, so part of the rise came from people who posted throughout. Their edit gradient has the same sign as the site-level one, but its placebo is significant and of opposite sign, so I do not rely on it.

Deleted answers are missing from the dump, and under the rules a suspected model answer was grounds for deletion. The Internet Archive holds a copy of the September 2023 dump, and matching its 3.55 million panel answers from January 2019 to August 2023 to the April 2024 dump identifies the answers deleted in between. Only 4{,}663 disappeared, 1{,}867 of them posted after the release. On code-heavy sites the deleted post-release answers had nearly twice the marker rate of those that stayed, but they were too few to move any estimate, and the code and edit coefficients on the September copy match those on the April dump. Post-release deletion rose less on heavily edited sites, which is the opposite of what deletion would need to produce the gradient. The September copy cannot show deletions made before it was taken. As a proxy I use the share of post and answer ids in each month that are missing from the dump. From December 2022 to August 2023 this share fell more on heavily edited sites than on lightly edited ones, though not significantly, so I find no sign that heavily edited sites deleted more in early 2023.

\section{Discovery Fails on a Mixed Panel}

Following \citet{kobak2025}, I also let the data choose the words. I split the sites into two halves stratified by code exposure, fitted a linear trend to each word's document share from 2021Q1 to 2022Q3 on one half, and kept the 100 words whose share from 2023Q2 to 2024Q1 most exceeded the extrapolation. One half's list begins \textit{configuration, overall, skills, children, sides, mainly, growing, marriage}, and the other has \textit{oxygen} and \textit{egypt} among its first 20. A few model words appear further down. Scored on the held-out half, neither list shows a dose response, and one has a significant negative placebo.

Discovery works on biomedical abstracts because the corpus is enormous and its topics change slowly relative to the style signal. A panel of 170 sites has 170 topical drifts that do not cancel over 85 sites, and the procedure returns whichever topics grew. I do not recommend data-driven discovery for panels of this kind. A list chosen on other text misses vocabulary specific to the platform, but combined with frequency-matched control lists it gave stable estimates here.

\section{Discussion}

I argue that the gradient is in the text as posted and follows how often authors revised their own answers. It is steeper among answers that were never edited, and late edits left the marker count almost unchanged, so it does not depend on words removed afterwards. It follows the self-edit share, and the share of answers edited by other users adds little once the self-edit share is held fixed. This is evidence against one form of a scrutiny account, in which users avoid posting model text where others will correct it, although self-edits may themselves respond to comments by others. Among the accounts that remain, habitual self-editing may mark authors who write their own answers and keep them up to date, and such authors may have had less use for generated text. The self-edit share may also mark a type of site, since the least edited sites are mostly small product-support and technology forums, where answers are written once to solve a specific problem, and model answers may have been both more useful and more tolerated there. Finally, sites with more editing may have detected and deleted more model answers before September 2023. The id-gap proxy and later deletion both point the other way, but neither rules this account out.

Frequency-based estimates of model use \citep{liang2024monitoring,liang2025widespread,kobak2025} infer adoption from the vocabulary that appears in a corpus. On Stack Exchange the marker rise ranged from about 29 percent on the least edited third of sites to none on the most edited third. I have no data on use, and the first two accounts above imply that the vocabulary may track use accurately, with less use where authors edit their own work. The data show that the marker rise varied this much across sites and that it was correlated with editing habits. They do not show that editing changed what the vocabulary records for a given level of use. Testing whether site practice also affects the measurement requires data on use, such as surveys or usage logs, site by site.

Many sites in the network announced rules against generated answers within weeks of the release, and this paper does not code which sites did so or when. During this period the marker rise ranged from none to about 29 percent across thirds of sites. Without sites known to have had no rule, the data say nothing about what the rules did, and the variation may come from how useful the model was, who was posting or how much demand there was for answers.

\citet{kabir2024} found that about half of ChatGPT's answers to Stack Overflow questions contained incorrect information. Marker words do not identify generated answers one by one, and I measure vocabulary and not accuracy, so whether the concentration on lightly edited sites affected the accuracy of the archive is open. Rating the accuracy of 2023 answers on the least and the most edited sites, and recording how often wrong ones were later corrected, would test it.

Experiments find that writing with a language model makes texts more alike \citep{padmakumar2024,doshi2024}. The shift of regular authors toward the marker words suggests that this convergence can appear in the writing of people who kept posting on Stack Exchange after the ban. I cannot tell whether these authors pasted model text, revised it, or adopted the words from reading model output, and the author-level estimates are less precise than the aggregate ones.

\paragraph{Limitations.} I did not plan the edit-share analysis. It came out of a screen of 15 traits, and although the placebo window, the split halves and the alternative measures argue against chance, replication on other platforms is needed. I do not observe use of the model, so the gradient cannot distinguish lower use on sites with a high self-edit share from any effect of editing on what the vocabulary records. The control lists are matched only on frequency, and the register lists address only the most obvious alternative to a model signal. All estimates compare sites with one another, and the month effects absorb any change common to the network. The marker list was built on scientific text, and a list built on Stack Exchange might be more sensitive. Much of the gradient rests on \textit{ensure}. The dump holds only the final version of each answer, so pre-period scores are measured on final text, edits made within the five-minute grace period are not recorded, and I cannot score the text as first posted. I cannot observe drafts, flags or deletions before September 2023 except through the id-gap proxy. The split into self- and other-edits identifies who edited an answer and says nothing about who read it.

\paragraph{Future directions.} Stack Exchange dumps are a common source of training data for language models, and models trained on recursively generated text lose the tails of the original distribution \citep{shumailov2024}. I hypothesize that model-written text enters such data unevenly, concentrated on lightly edited sites, so that the share of generated text in a scraped corpus differs across its sources with site practice. Testing this needs measures of model output validated within each site and a comparison of platforms that ban generated content with platforms that allow it. Separating the three accounts above is a second task, for which the timing of an author's self-edits around their first answer containing marker words is one source of evidence.

\section{Conclusion}

Stack Overflow banned answers generated by ChatGPT in December 2022, and many other Stack Exchange sites announced similar rules. I argue that vocabulary-based estimates of model use on platforms like this one should be reported site by site. In 2023 the vocabulary that language models overuse rose about 14 percent above trend across 170 sites and about 29 percent on the third of sites whose answers were least often edited before the release, with \textit{ensure}, \textit{essential} and \textit{furthermore} among the largest contributors, and on the most edited third it did not change. Without data on use, I cannot say whether sites with a high self-edit share used the model less or whether their practices changed what the vocabulary records. The gradient follows the self-edit share, and the other-edit share adds little once the self-edit share is held fixed.

\appendix
\section{Appendix}

Table~\ref{tab:code} reports the code-exposure specifications, the list and header outcomes, the math falsification dose and the regular-author estimates discussed in the Results.

\begin{table}[h]
\centering
\small
\setlength{\tabcolsep}{4pt}
\begin{tabular}{@{}lcc@{}}
\toprule
 & ChatGPT & Placebo \\
\midrule
\multicolumn{3}{l}{\textit{Marker rate on code rank}} \\
\quad all sites & 0.0082 (.002) & 0.0014 (.52) \\
\quad with size, length, math & 0.0033 (.15) & \\
\quad no 10 most exposed & 0.0119 (.002) & \\
\quad no 10 largest & 0.0105 (.005) & \\
\quad unweighted & 0.0074 (.17) & \\
\quad $\geq$50 prose tokens & 0.0118 ($<$.001) & \\
\quad no lists or headers & 0.0084 (.003) & \\
\addlinespace
\multicolumn{3}{l}{\textit{Other outcomes on code rank}} \\
\quad share with a list & 0.0231 (.003) & 0.0012 (.87) \\
\quad share with a header & 0.0189 (.039) & $-$0.0009 (.90) \\
\quad mean log words & 0.066 (.056) & $-$0.018 (.46) \\
\quad log words, no trend, 2021-- & 0.012 (.78) & $-$0.066 (.023) \\
\addlinespace
\multicolumn{3}{l}{\textit{Marker rate on math rank}} \\
\quad alone & $-$0.0073 ($<$.001) & \\
\quad with size, length, edit & $-$0.0014 (.63) & \\
\addlinespace
\multicolumn{3}{l}{\textit{Regular authors, code rank}} \\
\quad marker rate & 0.0051 (.22) & 0.0001 \\
\quad any marker & 0.0092 (.035) & \\
\bottomrule
\end{tabular}
\caption{Code exposure, other outcomes and regular authors. Same specification as Table~\ref{tab:panel}, with site trends except in the row marked no trend. Regular authors are the 9{,}573 with at least three answers in total in the 12 months before the release and at least three in the 12 months after. Placebo entries are left blank where the check was not run.}
\label{tab:code}
\end{table}

\bibliography{refs}
\end{document}